\documentclass[10pt]{article}
\usepackage{amsfonts}
\usepackage{epsfig}

\usepackage{cite}

\usepackage{amsmath}

\usepackage{color}

\begin{document}

\title{\LARGE \bf Evolution of the lexicon: a probabilistic point of view}

\author{Maurizio Serva}
\maketitle
\centerline{\it Dipartimento di Ingegneria e Scienze dell'Informazione 
e Matematica, 
Universit\`a dell'Aquila, L'Aquila, Italy.}

\begin{abstract}
The Swadesh approach for determining the temporal separation between two languages 
relies on the stochastic process of words replacement
(when a complete new word emerges to represent a given concept).
It is well known that the basic assumptions of the Swadesh approach are often unrealistic
due to various contamination phenomena and misjudgments (horizontal transfers,
variations over time and space of the replacement rate, 
incorrect assessments of cognacy relationships, presence of synonyms, and so on).
All of this means that the results cannot be completely correct.

More importantly, even in the unrealistic case that all basic assumptions are satisfied, simple 
mathematics places limits on the accuracy of estimating the temporal separation
between two languages.
These limits, which are purely probabilistic in nature and which are often neglected in 
lexicostatistical studies, are analyzed in detail in this article.

Furthermore, in this work we highlight that the evolution of a language's lexicon is also driven
by another stochastic process: gradual lexical modification of words. 
We show that this process equally also represents a major contribution
to the reshaping of the vocabulary of languages
over the centuries and we also show, from a purely probabilistic perspective, that taking
into account this second random process significantly increases the precision in determining 
the temporal separation between two languages.

\end{abstract}

\begin{flushleft}

\end{flushleft}

\section{An unavoidable preamble}
Glottochronology is a lexicostatistical tool that allows us to establish when a mother tongue gave 
rise to two daughter languages, or, what is the same thing, how many years have passed since 
the most recent common ancestor of two related languages.
This methodology, in principle, also allows us to reconstruct the family trees of language families, 
with dates assigned to each branching point.

The theory, conceived by Morris Swadesh in the 1950s, was inspired by the success 
of the carbon-14 dating technique developed at the time
\cite{Swadesh:1950,Swadesh:1951,Swadesh:1952,Swadesh:1954,Swadesh:1955}.

Swadesh's founding hypothesis was that a word associated with a given concept in 
a language could be replaced in time by a totally new one at a constant probability rate, 
exactly as a carbon-14 atom can be replaced at a constant rate by a more stable carbon atom.

Actually, the replacement of a word by a totally new synonym in a language is a rare but observable 
event whose probability rate seems to be more or less constant along the centuries.
Therefore, if one considers the $M$ words corresponding to
$M$ concepts in a given language (a Swadesh list),  the number of un-replaced words 
after a time $T$ will be approximately 
\begin{equation}
\mathcal{M}(T) \simeq M {e}^{ - \lambda T},
\label{sw1}
\end{equation}
where $\lambda$ is the replacement rate of a single word.
Obviously, $\mathcal{M}(0)=M$ (there are no changed words at initial time)
and $\lim_{T\to \infty} \mathcal{M}(T)=0$ (all words are changed at very large times).
The replacement rate was estimated by Swadesh to be $\lambda =1.4 \times10^{-4}$ per year,
in  Appendix B we give an independent estimate of $\lambda$ which confirms this value.

The consequence of this assumption can be easily understood by an example:
consider the same Swadesh list of $M$ concepts in Spanish and Latin and suppose you find by a 
direct check a number $\mathcal{M}$ of pairs of words in the two languages which correspond to 
the same concept and which are cognates (equal words in this context since eventual difference
is only due to random modification in spelling and pronounciation due to geographical, political and 
historical factors), then your estimate of the separation time $T$ between the two languages
can be obtained by means of (\ref{sw1}) as
\begin{equation}
T \simeq  -\frac{1}{\lambda} \ln \! \left(\! \frac{\mathcal{M}}{M} \!\right).
\label{sw2}
\end{equation}

More often it happens you don't know the vocabulary of of the ancestor language, 
nevertheless you would like to know the time separation between two contemporary 
languages, which means the time from their most recent common ancestor.  
In this case, the Swadesh assumption tells you that the probability rate
that in the pair of words corresponding to the same concept in the two languages
at least one of them is replaced is $2 \lambda$.
Therefore, given  two Swadesh list of $M$ words, the number of cognate pairs 
in the two languages after a time $T$ will be  approximately
\begin{equation}
\mathit{\Omega}(T) \simeq M {e}^{ - 2\lambda T},
\label{sw3}
\end{equation}

The main consequence of this hypothesis can be easily understood with an example: consider 
the same Swadesh list of $M$ concepts in Spanish and Italian
and suppose to find by direct observation 
a number $\mathit{\Omega}$ of pairs of words in the two languages that correspond to the same
concept and that are cognate (equal words in this context since any 
eventual difference is only due to small repeated random changes in phonology and/or orthography), 
then, the estimated separation time 
$T$ between the two languages can be obtained by means of (\ref{sw1}) as
\begin{equation}
T \simeq -\frac{1}{2\lambda} \ln \! \left( \! \frac{\mathit{\Omega}}{M} \! \right).
\label{sw4}
\end{equation}
Hereafter we will name "Cognate Overlap" the intensive variable 
$\omega=\frac{\mathit{\Omega}}{M}$ 
which goes from 0 (languages totally different) to 1 (coinciding languages).
This is the glottochronological formula, as first proposed by Swadesh in
\cite{Swadesh:1950} (as an aside, in a footnote he gives credits to the physicist 
Gordon Gould for the mathematical aspects of the work).
This technique is the lexicostatistical equivalent of the molecular clock, used in evolutionary 
biology to estimate the time elapsed since two species diverged from their common ancestor.  
The molecular clock, similarly to glottochronology, is based on the idea that DNA and protein 
sequences evolve at a constant rate over time and it is therefore based on counting the number 
of genetic coincidences between two species.

There is a number of reasons why the equalities  (\ref{sw2}) and (\ref{sw4}) 
do not exactly hold, the main of which are: 
\begin{itemize}
\item
the hypothesis the rate of replacement remains perfectly constant in time may be not respected. 
Evidence is that the rate depends on historical events, for example at the end of
Roman Empire, there were rapid lexical evolution due to the political and social changes
caused by the entry of new populations in regions previously under Roman rule, on the contrary, 
the later standardization of national Romance languages was a factor in slowing down 
their lexical evolution; 
\item
the creation of a Swadesh list implies a choice between 
various synonyms for any concept. This choice is not univocally determined by a codified
rule and, in consequence, the cognate overlap may change according to different approaches.
For example in Italian there are two different words to say cheese, one is $formaggio$, which 
is the same as the French $fromage$, the other is $cacio$, the same as
the Spanish $queso$; 
\item
languages swap words, thus increasing their cognate overlap (if loans are not detected).
For example, Italian borrowed the following words from French: toilette, bricolage and bidet
and both languages borrowed some common words from English.
This phenomenon, named horizontal transfer,  is typically more relevant between 
those languages that are geographically closer;
\item
there is a big problem concerning the individuation 
of cognates ($i.e.$ equal words). It is easy to see that the Italian word $uomo$ 
and the French word $homme$ are the same, it is less evident that
$leite$ in Portuguese and $\gamma \acute\alpha \lambda \alpha$ ($g \acute a la $) 
in Greek are the same word, we can only infer this through
historical phonological reconstruction. 
Moreover, the detection of cognates is a matter of individual choice and, therefore, it
may be different for different scholars.
\end{itemize}

The point of this paper is that there is a more fundamental problem with
the  equalities (\ref{sw1}) and (\ref{sw3}):  $\mathcal{M}(T)$ and 
$\mathit{\Omega}(T)$ are random quantities
so that statistical fluctuations from the average may lead to an erroneous evaluation of the 
separation time by  (\ref{sw2}).
In fact, the standard deviation of  $ \frac{\mathcal{M}(T)}{{M}}$ and   
$ \omega= \frac{\mathit{\Omega}(T)}{{M}}$ 
are proportional to $ \frac 1 {\sqrt{M}}$. While  fluctuations are not a problem 
concerning the carbon-14 decay since $M$
(the number of atoms) is of the order of the Avogadro number ($ M \simeq 6 \times 10^{23}$),  
the largest Swadesh list only contains $M=207$ concepts.
Therefore,  the equality (\ref{sw1})  only holds on average while
the observed values of 
the stochastic variable $\mathcal M (T) $ can be well different from its average, as we will show.
Exactly the same problem arise with the equality (\ref{sw3}) for the
stochastic variable $\mathit{\Omega}(T)$.

This is a point that is widely underestimated and/or misunderstood in lexicostatistics: 
mathematics alone sets a limit to the precision with which one can 
determine the separation time between two languages.
In this paper we only focus on this probabilistic limits
neglecting the other relevant sources of possible incorrect estimation listed above.
Therefore we assume that the scholar dealing with Swadesh approach is 
in the ideal, although highly unrealistic, situation in which:
\begin{itemize}
\item
the rate of replacement is strictly constant in time and it is universal
(the same for all languages); 
\item
he is able to univocally draw up the Swadesh list  
(absence of synonyms);
\item
languages do not swap words (no horizontal transfers are allowed);
\item
 he is able to determine cognacy with absolute precision 
(no false cognate pairs, no missing cognate pairs).
\end{itemize}

Nevertheless, as we will show, even if these unrealistic ideal assumptions are satisfied,
the estimated and the real separation times
can be catastrophically different only because of probabilistic fluctuations.
This is really a relevant bug which is to be attributed exclusively to the fact that the number of 
concepts is small (207) at variance with carbon-14 dating where $M$, the number of atoms,
is thermodynamically large  ($6 \times 10^{23}$).
We will perform in Section 2 a precise analysis on the mathematical limits of validity of the 
Swadesh method assuming that all the ideal conditions listed above are satisfied.
  
A second, but not less relevant, point of this paper is a mathematical 
analysis concerning the advantage of using a different and, ultimately, complementary strategy, 
to evaluate the time separation between languages.
Swadesh only used lexical replacements, while he didn't consider the gradual independent 
modification of cognate words as a possible source of information.
The idea to take advantage of this second source arises from the simple observation that 
two initially identical terms employed in different geographical locations (say Madrid and Florence)
become gradually different because of small cumulative modifications, even in absence 
of replacements (such as $homo$ $\to$ $hombre$/$uomo$).

Although a single random change in one of two cognate words (such as the modification of a single 
character) has a smaller impact concerning their differentiation compared to a replacement, 
it is much more frequent. 
We will show, indeed, that the processes of replacement and gradual modification 
have comparable cumulative  random effect on lexicon evolution, but the second with smaller fluctuations.
In sum, replacements and gradual modifications
equally contribute to the reshaping of the vocabulary of languages over the 
centuries, but taking into account the second random process significantly increases the precision 
in determining the time separation between two languages.

As for replacements, we have here a purely probabilistic perspective described in Section 3, 
which assumes that random gradual modifications take place at a constant and universal rate $\mu$.
Moreover we will make the simplifying assumptions that all words are composed by the same 
number $L$ of characters (the length of words is $L$) and that each character may assume $N$ 
possible values (the effective number of letters of the alphabet is $N$).
Both the effective quantities $L$ and $N$ are experimentally estimated in Appendix B.

In Section 4 we will describe a strategy where only the "Normalized Hamming Distance" of 
words is used to compare the difference between languages. In our probabilistic approach we 
maintain the simplification that all words have equal length.
The method applies blindly to all concepts, independently of the fact that the two associates words 
in the two languages are  cognate or not. Therefore, with respect to the Swadesh approach, the
advantage is that no prior identification of cognates is requested, thus avoiding 
all possible human errors associated with this procedure.
On an experimental ground, the method was proposed in \cite{Serva:2008,Petroni:2008} 
and impressively later applied to the huge Automated Similarity Judgment Program
(ASJP) dataset, which includes nearly all recognized language families in the 
world and many subfamilies \cite{Holman:2011}.
In these researches, given that real words may have different length, 
"Normalized Levenshtein Distance"  was used instead of normalized Hamming distance
In fact, while Levenshtein \cite{Levenshtein:1966} and Hamming distances are both edit
distances the first also allows for insertions and deletions and for this
reason it has to be used in case of words of different length.

More recently a combined strategy has been used where 
word distances have been calculated only for cognate pairs \cite{Pasquini:2023}. 
In Section 5 we analyze on a mathematical ground
this strategy, which, for very large separation times, works better than the the 
blind strategy of Section 4, nevertheless, the price to pay is the possible errors associated 
with the identification of cognate terms, as in the Swadesh approach.

\section{Cognate overlap: the classical glottochronology}
The number of concepts entering in a Swadesh list is $M$, each of them is individuated by
an index $i=1, \dots M$.
Given a pair of contemporary languages
one has $M$ words for the $M$ concepts for each of them.
Thus, we define $M$ independent stochastic variables $\sigma_i (t) $  
such that $\sigma_i(t) \! =1$ if the two words for the concept $i$
are cognates, otherwise $\sigma_i(t) \! =0$. 
Therefore, the number of cognate pairs at time $T$  (time from the last common ancestor)
is 
\begin{equation}
\mathit{\Omega}(T) = \sum_{i=1}^M \sigma_i (T).
\label{sua1}
\end{equation}
The cognate overlap, which we have defined as
\begin{equation}
 \omega(T)=\frac{\mathit{\Omega}(T)}{M},
 \label{sua2}
\end{equation}
goes from 0 (totally different languages) to 1
(coinciding languages). 

According to Swadesh the probability that at least one of the two words in a cognate pair
is replaced in a time $dt$ is $2\lambda dt$.
When one of the two words is replaced a pair of cognates becomes a pair of non-cognates
while a pair of non cognates remains  a pair of non-cognates. 
Therefore, if $\sigma_i(t) \! =1$ it may vanish at time $t+dt$ with probability $2\lambda dt$ 
while if $\sigma_i(t) \! =0$ it remains unchanged at time $t+dt$ with probability 1.
One immediately has the differential equation for the expected value
at the left below
\begin{equation}
\frac{d}{dt}{\rm E}[\sigma_i(t)] = -2 \lambda {\rm E} [\sigma_i(t)]
\,\,\,\,\,\, \to \,\,\,\,\,\,
{\rm E} [\sigma_i(T)] = e^{-2 \lambda T},
\label{sua3}
\end{equation}
where the solution at the right above is obtained by assuming
that  all $\sigma_i(0)$ equal 1 (at  time $T=0$ the two languages 
start to differentiate).
Given that $\sigma_i (T) $ is a Bernoulli variable 
which takes the values 0 and 1, the probability that $\sigma_{i} (T)=1$ equals 
the expected value ${\rm E} [\sigma_{i} (T)]$, while the probability that $\sigma_{i} (T)=0$ equals 
$1- {\rm E} [\sigma_{i} (T)]$, therefore,
\begin{equation}
\sigma_{i} (T) = 
\begin{cases}
1 \,\,\,\,\,\,\,\,\,  {\rm with \; probability}  \,\,\,\,\,\, \,\,\,  e^{-2 \lambda T} 
\\[0.6 ex]
0 \,\,\,\,\,\,\,\,\,  {\rm with \; probability}   \,\,\,\,\,\, \,\,\, 1-e^{-2 \lambda T}.
\end{cases} 
\label{sua4}
\end{equation}

The variable $\sigma_i(T)$ can be interpreted as the cognate overlap between the two words
corresponding to the same concept $i$ while $1- \sigma_i(T)$ is their cognate distance
(if $\sigma_i(T)=1$ the distance equals zero, if $\sigma_i(T)=0$ the distance equals one).
The cognate distance between the two languages 
can be simply defined as $\frac 1 M \sum_i (1-\sigma_i)= 1-\omega(T)$,
which ranges from 0 to 1 and equals zero for two identical languages
(all pairs are composed by cognate words)
and equals one when the two languages are completely different
(all pairs are composed by non-cognate words).
In the last case one concludes that the two languages have no common ancestor.

Given that the variables $\sigma_i(T)$  only take the values 0 and 1, one 
has $\sigma_i^2(T)=\sigma_i(T)$ 
and given the expectation at the right in (\ref{sua3}) it follows 
$ {\rm E}[\sigma_{i}^2 (T) ]= {\rm E}[\sigma_{i} (T) ] = e^{-2\lambda T} $,
which implies the variance
\begin{equation}
{\rm V\!ar} \! \left[\sigma_{i} (T) \right]= 
{\rm E}[\sigma_{i}^2 (T) ]-
{\rm E}[\sigma_{i} (T) ]^2 
= e^{-2\lambda T} (1-e^{-2\lambda T}) .
\label{sua5}
\end{equation}

According to (\ref{sua1}) and (\ref{sua4}) , 
the variable $\mathit{\Omega}(T)$ follows a binomial distribution of 
parameters $e^{-2 \lambda T}$ and  $M$.
We  prefer to compute here directly its expected value and its variance 
since we will late use similar calculations for variables slighter more complicated.
Given that $\mathit{\Omega}(T)$  is the sum of $M$ independent and identically 
distributed variables and given (\ref{sua2}), not only one has that
\begin{equation}
{\rm E} \! \left[  \omega(T)    \right] =  {\rm E}[\sigma_i(T)] = e^{-2 \lambda T},
\,\,\,\,\,\,\,\,\,\, 
{\rm V\!ar} \! \left[  \omega(T)    \right]  = \frac 1 M {\rm V\!ar} \! \left[\sigma_{i} (T) \right]
=\frac 1 M e^{-2\lambda T} (1-e^{-2\lambda T}), 
\label{sub1}
\end{equation}
but also that $\omega(T)$ is approximately gaussian distributed, therefore,
with probability $95\%$:
\begin{equation}
{\rm E} \! \left[  \omega    \right]  - 2\sqrt{{\rm V\!ar} \! \left[  \omega   \right]  }  
\le \omega  \le
{\rm E} \!\left[  \omega    \right] + 2\sqrt{{\rm V\!ar} \! \left[  \omega    \right]  }  
\label{sub2}
\end{equation}
where here and hereafter the argument $T$ is dropped.

\begin{figure}
\centering
\includegraphics[height=9cm, width=15cm,angle=0]{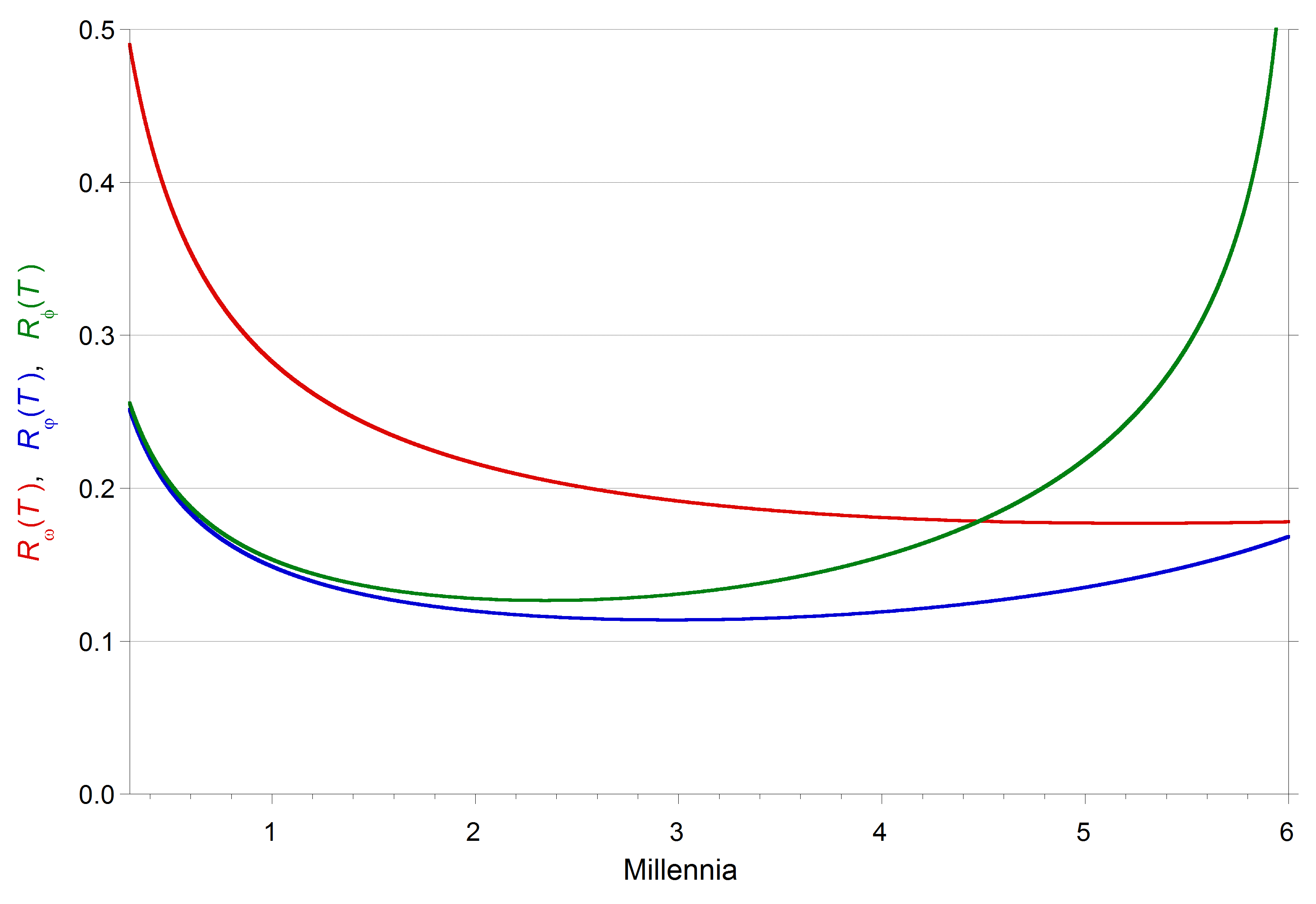}
\caption{The relative errors $R_\omega$ (red) as defined by (\ref{sub7}), 
concerning the classical Swadesh approach,
$R_\phi$ (green) as defined by (\ref{leb1}), (\ref{leb6}) and (\ref{A15}),
concerning the blind use of normalized edit distance
and $R_\varphi$ (blue) as defined by (\ref{me3}), (\ref{me6}) and (\ref{A11}),
concerning the combined use of edit distance and cognate identification. 
We plot the three errors in function of $T$ in the interval $300 \le T \le 6,000$. 
We use the Swadesh proposal for the replacement rate $\lambda =1.4 \times10^{-4}$
which is confirmed by our estimate. We assume $M=207$, again according to Swadesh.
Moreover, we use our estimates $\mu=1.6 \times10^{-4}$, $N=5.18$ and $L=7.63$.}
\label{rerr}
\end{figure}

From the first equality in (\ref{sub1}) one trivially derives
\begin{equation}
T = -\frac{1}{2\lambda} \ln \! \big( {\rm E} \! \left[ \omega \right] \!  \big),
\label{sub3}
\end{equation}
nevertheless, the observed stochastic separation time $ \mathcal{T}_{\omega}$ is
\begin{equation}
 \mathcal{T}_{\omega}= -\frac{1}{2\lambda} \ln \! \left(\omega \right).
 \label{sub4}
\end{equation}
It is the second quantity which forcefully enters in Swadesh method
(a single realization of the stochastic processes) as well  in its
subsequent versions (see, for example,\cite{Dyen:1967,Starostin:1989,Starostin:2010})
and in all applications (see, for example,  \cite{Verin:1969,Gray:2000,Gray:2003,Greenhill:2009}).
So the point is how different it is $ \mathcal{T}_{\omega}$ from $T$? 
According to (\ref{sub2}) one has with probability $95\%$
\begin{equation}
T_-  \le \mathcal{T}_{\omega} \le  T_+
\label{sub5}
\end{equation}
where 
\begin{equation}
T_\pm =-\frac{1}{2\lambda} \ln \! \left( {\rm E} \! \left[ \omega \right] \mp 
2\sqrt{{\rm V\!ar} \! \left[  \omega    \right] } \right)
\label{sub6}
\end{equation}
A reasonable measure of the relative error on separation time is therefore
\begin{equation}
R_\omega =  \frac{T_+ \!  - \! T_-}{2T} =
\frac{1}{4\lambda T} \ln \! 
\left( \frac{ {\rm E} \! \left[ \omega \right] + 2\sqrt{{\rm V\!ar} \! \left[  \omega    \right] } }
{ {\rm E} \! \left[ \omega \right] - 2\sqrt{{\rm V\!ar} \! \left[  \omega    \right] } } \right) = 
\frac{1}{4\lambda T} \ln \! 
\left( \frac{1 + 2\sqrt { \frac{e^{ 2 \lambda T} -1}{ M}}}{1 - 2\sqrt { \frac{e^{ 2 \lambda T} -1}{ M}}}  \right).
 \label{sub7}
\end{equation}
It should be noticed that when $M$ 
refers to the number of atoms in a decay process 
this error disappears since $M$ is of the order of the Avogadro number
($ M \simeq 6 \times 10^{23}$), but it is relevant for glottochronology  where 
$M=207$ as it can be seen in Fig. (\ref{rerr}) 
where $R_\omega$ is plotted as a function of $T$ for $3 \times 10^2 \leq T \leq 6 \times 10^3$. 
In fact $R_\omega$ ranges from $49\%$ for very short times (0.3 millennia)
to $18\%$ for large times (6 millennia). 
This error also implies that one hardly can reconstruct the exact topology
of the genealogical tree of a family of languages, especially if the ancestor language,
is to close to the present  (as it is for the Romance family)
or there is more than a single branching event close to the root.

\section{ Normalized Hamming distance and overlap}

It is a matter of fact that words, even in absence of a replacement, are subject to modifications 
over time. This explain why the Latin word $caseum$ transformed into the cognate words 
$queso$ in Castillan and $cacio$ in Italian.
We assume that these random gradual modifications take place at a constant and universal rate 
$\mu$, which is the mirror hypothesis of a constant $\lambda$.
Moreover, we make the simplifying assumptions that all words are composed of the same 
number $L$ of characters (the length of words is $L$) and that each character may assume $N$ 
possible values (the effective number of letters of the alphabet is $N$).
Both the effective quantities $L$ and $N$ will be experimentally estimated in Appendix B
comparing pairs of words that certainly have no common ancestor.

These values may significantly differ from the nominal values (the number of letters 
in the English alphabet is 26 but $N$ typically has a much smaller value).
The reasons for this discrepancy are fundamentally two; the different
frequency of different  letters and the fact that letters are not monads (in english 
$h$ appears often after a $t$).
Nevertheless, if we replace our estimates for $N$ and $L$ respectively by 26
(letters in the english alphabet) and by the average length of english words, our final argument 
would be even strengthened. 
Indeed, the real values of $N$ and $L$ are not truly relevant for our argument.

The normalized Hamming overlap of two words is simply the number of pairs of 
equal characters (in the same position) in the two words, divided by the length 
$L$ of the words. This number ranges from 0 (totally different words) to 1 (identical words). 
The normalized Hamming distance of two words, as defined in standard way,  is simply the number 
of pairs of different characters (in the same position) in the two words, divided by $L$.

It is very important to remark that from an operative point of view 
there is no need of deciding if the two words corresponding to the same concept $i$ 
are cognate or not, nevertheless, the hidden probabilistic structure makes a difference between 
pairs of cognate words (whose normalized Hamming overlap is $ \eta_i(T)$) and
pairs of non-cognate words (whose overlap is $ \xi_i(T)$).
In fact, two non cognate words have no common origin
and, therefore, two characters in the same position may be equal
at any time $T$ only by mere chance. On the contrary, 
for two cognate words the two characters are necessarily equal at initial time $T=0$
(words are initially identical).

Let us now consider the case of a pair of non-cognate words corresponding to the 
same concept $i$, 
the letter in position $k$ is the same for the two words only by chance 
(at any time $T$, replacements of letters or words do no not alter this fact!)
therefore having defined $\xi_{i,k}(T)$ as the variable which takes the value 1 if the two letters
are equal and 0 otherwise, one has 
\begin{equation}
\xi_{i,k} (T)= 
\begin{cases}
1 \,\,\,\,\,\,\,\,\,  {\rm with \; probability}  \,\,\,\,\,\, \,\,\,  \frac 1 N
\\[1.2 ex]
0 \,\,\,\,\,\,\,\,\,  {\rm with \; probability}  \,\,\,\,\,\, \,\,\, \frac{N\!\!-\!\!1}{N}.
\end{cases} 
\label{sc1}
\end{equation}
Then one can define the two variables
\begin{equation}
\xi_i (T) = \frac{1}{L} \sum_{k=1}^L  \xi_{i,k}(T),
\,\,\,\,\,\,\,\,\,\,\,\,\,\,\,\,\,\,\,\,
1-\xi_i(T)  = \frac{1}{L} \sum_{k=1}^L  (1- \xi_{i,k}(T)),
\label{sc2}
\end{equation}
which are respectively the normalized Hamming overlap and 
the normalized 	Hamming distance for two non-cognate words 
corresponding to the concept $i$.

Let us now consider the case of a pair of cognate words corresponding to the 
same concept $i$,  the two letters in position $k$ are  the same at initial time but later this fact 
may change because one of the two letters can be randomly substituted by a new one.
We assume that in a time $dt$ a letter value can be substituted  with probability $\mu dt$
by one of the  $N$ possible values (thus, including the departure value), 
therefore having defined $\eta_{i,k}(t)$ as the variable which takes the value 1 if the two letters
are equal and 0 otherwise, one has that if $\eta_{i,k} (t) \! =1$ then $\eta_{i,k} (t+dt) \! $ 
may vanish at time $t+dt$ 
with probability $2\mu \frac{N\!\!-\!\!1}{N} dt$,  while if $\eta_{i,k} (t) \! =0$  then
 $\eta_{i,k} (t+dt) \! $ may turn to the value 1 at time $t+dt$ with probability 
 $ \frac{2 \mu}{N}dt$.  In a compact form:
\begin{equation}
\eta_{i,k} (t+dt ) = 
\begin{cases}
1- \eta_{i,k} (t) \,\,\,\,\,\,\,\,\,  {\rm with \; probability}  \,\,\,\,\,\, \,\,\,
2 \mu \frac{1+ (N-2)\eta_{i,k} (t)}{N} dt  \\[1.2 ex]
\eta_{i,k} (t) \,\,\,\,\,\,\,\,\,  {\rm with \; probability}  \,\,\,\,\,\, \,\,\,
1-2 \mu \frac{1+ (N-2)\eta_{i,k} (t)}{N} dt. \\[0.3 ex]
\end{cases} 
\label{sc3}
\end{equation}
After some biking, taking into account that $\eta_{i,k}^2(t)=\eta_{i,k}(t)$, one obtains the 
differential equation for the expected value at the left below
\begin{equation}
\frac{d}{dt} {\rm E} [\eta_{i,k} (t)] = -2 \mu \left({\rm E}[\eta_{i,k} (t)] - \frac 1 N \right)  
\,\,\,\,\,\, \to \,\,\,\,\,\,
{\rm E}[\eta_{i,k}(T)] = \left[ \frac{N\!\!-\!\!1}{N} e^{-2\mu T}+ \frac 1 N \right],
\label{sc4}
\end{equation}
where the solution at the right above is obtained by assuming
that  all $\eta_{i,k} (0)$ equal 1.

Given that $\eta_{i,k}(T)$ is a Bernoulli variable 
which takes the values 0 and 1, the probability that $\eta_{i,k}(T)=1$ equals 
the expected value ${\rm E}[\eta_{i,k}(T))]$, therefore,
\begin{equation}
\eta_{i,k}(T)= 
\begin{cases}
1 \,\,\,\,\,\,\,\,\,  {\rm with \; probability}  \,\,\,\,\,\, \,\,\,  
\left[ \frac{N\!\!-\!\!1}{N} e^{-2\mu T}+ \frac 1 N \right],
 \\[1.5 ex]
0 \,\,\,\,\,\,\,\,\,  {\rm with \; probability}   \,\,\,\,\,\, \,\,\, 
\frac{N\!\!-\!\!1}{N} \left[ 1-e^{-2\mu T}\right].
\end{cases} 
\label{sc5}
\end{equation}

Notice that the variables $\xi_{i,k}(T)$ also are subject  to the stochastic equations
(\ref{sc3}) but their probabilities does not vary in time since
they are in the steady state distribution already at $T=0$.
Notice, in fact, that the probabilities (\ref{sc1}) can be obtained from the (\ref{sc5}) in the limit 
$T \to \infty$ (when two languages lose memory of their common ancestor).

Let us stress that $\mu$ is the probability rate that a dice is rolled
independently of the output value which may eventually equal the departure one
with probability $\frac{1}{N}$, therefore
\begin{equation}
\hat \mu =\frac{N\!\!-\!\!1}{N} \mu
\label{sc6}
\end{equation}
is the probability rate that a letter in a word is substituted by a different one.
Our estimate for $\hat \mu$  is  $1.3 \times10^{-4}$ (see Appendix B). 
Since the probability rate of changing a given letter as a consequence of a replacement is
$\lambda=1.4 \times10^{-4}$, we conclude that the effect of replacements and gradual 
modifications have a comparable effect over lexical evolution.

Finally, one can define the two variables
\begin{equation}
\eta_i (T) = \frac{1}{L} \sum_{k=1}^L  \eta_{i,k}(T),
\,\,\,\,\,\,\,\,\,\,\,\,\,\,\,\,\,\,\,\,
1-\eta_i  (T)= \frac{1}{L} \sum_{k=1}^L  (1- \eta_{i,k}(T)),
\label{sc7}
\end{equation}
which are respectively the normalized Hamming overlap and 
the normalized 	Hamming distance for two cognate words 
corresponding to the concept $i$.

Expected values and variances of $\xi_i (T)$ and $\eta_i (T)$ are computed in Appendix A,
while $N$, $L$ and $\mu$ (the rate of a dice roll)
are experimentally estimated in Appendix B.
\section{Blind use of Normalized Hamming Overlap}

Suppose you are unable to decide if two words corresponding to the same concept in 
two different languages are cognate, or you simply want to avoid errors  
which may arise from this decision,
in this case you simply and blindly measure the normalized Hamming distance or overlap.
For what we have exposed in the two previous sections,
It is immediate to find out that this overlap is
\begin{equation}
\omega^H_i(T)=\sigma_i (T) \eta_i(T) + \big(1-\sigma_i(T)\big) \xi_i(T),
\label{lea1}
\end{equation}
because $\sigma_i(T)$ equals $1$ if the two words are cognate
and equals $0$ otherwise, moreover $\eta_i(T)$ apply in
the first case and $\xi_i (T)$ in the second.
The average and variance of the above overlap
are computed in appendix A.
Moreover, the normalized word distance according to Hamming is
\begin{equation}
d^H_i(T)= 1- \omega_i^H(T) =\sigma_i (T) \big(1- \eta_i(T) \big) +
 \big(1-\sigma_i(T)\big)\big(1- \xi_i (T)\big),
\label{lea2}
\end{equation}
whose average and variance are also computed in appendix A.

Let us stress once more that both overlap and distance, when operatively measured 
for a a pair of words of a dataset,
do not require that one knows if the two words are cognate or not,
but only a count of the number of pairs of equal (different) letters
for each of the $L$ positions.

It is also useful to define for future use
\begin{equation}
\phi_i(T) =  \frac{N}{N\!\!-\!\!1} \left[\omega^H_i (T)   -\frac{1}{N}  \right] 
= 1- \frac{N}{N\!\!-\!\!1} d_i^H (T) ,
\label{lea3}
\end{equation}
whose average and variance are derived in Appendix A.

The normalized Hamming overlap between two languages can be obtained by an average 
over all the concepts included in the Swadesh lists, therefore

\begin{equation}
\omega^H(T) = \frac 1 M  \sum_{i=1}^M\omega^H_i(T),
\label{lea4}
\end{equation}
which is a thermodynamically intensive variable.
Analogously, normalized Hamming distance between two languages is 
\begin{equation}
d^H (T)= \frac 1 M  \sum_{i=1}^M d^H_i(T) =1- \omega^H (T).
\label{lea5}
\end{equation}

It is worth to mention that on a experimental ground the method 
normalized Levenshtein distance replaces normalized Hamming distance 
\cite{Serva:2008,Petroni:2008,Bakker:2009,Petroni:2010,Wichmann:2010a,Wichmann:2010b, Holman:2011,Serva:2012a,Serva:2012b,Pasquini:2019,Serva:2020,Serva:2022,Wichmann:2023}, 
since words may have different lengths,  nevertheless
Hamming and Levenshtein are two different versions of edit distance
and, therefore, they are basically the same thing.
The reason why we use here Hamming distance is simply because 
the related mathematics is much easier.

It is finally also useful to define a third intensive variable as
\begin{equation}
\phi(T) =   \frac 1 M  \sum_{i=1}^M \phi_i(T) =
\frac{N}{N\!\!-\!\!1} \left[\omega^H(T)   -\frac{1}{N}  \right]
= 1- \frac{N}{N\!\!-\!\!1} d^H (T) .
\label{lea6}
\end{equation}
Given that $\phi(T)$ is the sum of $M$ independent and identically distributed variables
divided by $M$, one immediately has
\begin{equation}
{\rm E}\left[\phi(T) \right] ={\rm E}[\phi_i(T)] ,
\,\,\,\,\,\,\,\,\,\,\,\,\,\,\,\,\,\,\,
{\rm V\!ar} \! \left[\phi(T) \right] = \frac{1}{M} {\rm V\!ar} \! \left[\phi_i(T) \right],
\label{leb1}
\end{equation}
where ${\rm E}[\phi_i(T)] $ and ${\rm V\!ar} \! \left[\phi_i(T) \right]$, as already mentioned, 
are computed in Appendix A. It turns out
\begin{equation}
{\rm E}\left[\phi(T) \right] = e^{-2 (\lambda +\mu) T}
\label{leb1b}
\end{equation}
and
\begin{equation}
{\rm V\!ar} \left [\phi (T) \right] =  
\frac{1}{M}e^{-2\lambda T}(1-e^{-2\lambda T})e^{-4\mu T} + 
\frac{1}{ML }e^{-2\lambda T} (1-e^{-2\mu T}) \left[ e^{-2\mu T}+ \frac{1}{N\!\!-\!\!1}  \right]
+  \frac{1}{M L(N\!\!-\!\!1)}   (1-e^{-2\lambda T}).
 \label{leb1c}
\end{equation}

We can now follow the same line of reasoning of Section 2.
Since $\phi(T)$ is approximately gaussian distributed, with probability $95\%$ 
we have that
\begin{equation}
{\rm E} \! \left[  \phi    \right]  - 2\sqrt{{\rm V\!ar} \! \left[  \phi   \right]  }  
\le \phi \le
{\rm E} \!\left[  \phi   \right] + 2\sqrt{{\rm V\!ar} \! \left[  \phi    \right]  }  
\label{leb2}
\end{equation}
where here and hereafter the argument $T$ is dropped.
From the first equation in (\ref{leb1}) one trivially derives
\begin{equation}
T = -\frac{1}{2(\lambda + \mu)} \ln \! \big( {\rm E} \! \left[ \phi \right] \!  \big),
\label{leb3}
\end{equation}
nevertheless, the observed stochastic separation time $ \mathcal{T}_{\phi}$ is
\begin{equation}
 \mathcal{T}_{\phi}= -\frac{1}{2(\lambda+\mu)} \ln \! \left(\phi\right),
 \label{leb4}
\end{equation}
so the point is how different is it from $T$? According to (\ref{leb2}) one has that 
\begin{equation}
 -\frac{1}{2(\lambda+\mu)} \ln \! \left( {\rm E} \! \left[ \phi \right] 
+ 2\sqrt{{\rm V\!ar} \! \left[  \phi \right] } \right)\le \mathcal{T}_{\phi} \le
-\frac{1}{2(\lambda+\mu)} \ln \! \left( {\rm E} \! \left[ \phi \right] - 
2\sqrt{{\rm V\!ar} \! \left[  \phi \right] } \right) 
\label{leb5}
\end{equation}
with  probability $95\%$.
A reasonable measure of the relative error concerning separation time
can be found as in Section 2:
\begin{equation}
R_\phi =  
\frac{1}{4(\lambda + \mu) T} \ln \! 
\left( \frac{ {\rm E} \! \left[ \phi \right] + 2\sqrt{{\rm V\!ar} \! \left[  \phi   \right] } }
{ {\rm E} \! \left[ \phi \right] - 2\sqrt{{\rm V\!ar} \! \left[  \phi   \right] } } \right).
 \label{leb6}
\end{equation}
As it can be observed in Fig. 1,
the relative error $R_\phi$ is smaller than $R_\omega$ in the first four millennia,
this means that the edit distance approach should be considered 
as a valid alternative to the Swadesh approach in this interval of time.
This is even more true if we consider that there is no error caused by incorrect attributions
of cognacy.
The improvement is particularly relevant when the last common ancestor language 
is situated less then two millennia in the past (the relative error, in this case, is one half).
On the contrary, for large times the relative error $R_\phi$ explodes due to the increase 
of variance due to the contribution of non-cognate pairs, which, however, do not contain 
any useful information regarding the time $T$.

\section{Normalized Hamming Overlap limited to cognate pairs}

Once we discovered that measuring overlap of words in place of the number of cognates
leads to useful results,  we can try to combine the two strategies
as it was hypothesized in \cite{Pasquini:2023}.
We construct, therefore, a new intensive variable $\varphi$ which only uses the overlap
$ \eta_i(T)$ of cognate words and which takes the place of $\omega$
(Swadesh approach) or $\phi$ (blind use of Hamming distance).
Let us preliminary define
\begin{equation}
\varphi_i(T) =  \frac{N}{N\!\!-\!\!1} \sigma_i(T)  \left[ \eta_i(T)   - \frac 1 N \right],
\label{me1}
\end{equation}
which vanishes when the two words corresponding to the concept $i$ are 
non-cognates (when $\sigma_i(T)=0$) and which linearly depends on $\eta_i(T)$
when they are non-cognate (when $\sigma_i(T)=1$).
The average and variance of this variable are computed in Appendix A.

Then, it is straightforward to define the intensive variable
\begin{equation}
\varphi(T) = \frac 1 M  \sum_{i=1}^M \varphi_i(T).
\label{me2}
\end{equation}
It is worth noting that only cognate pairs contribute to this sum, therefore, from an operational 
point of view, all concepts that have cognate words must be identified and the Hamming 
distance must be calculated only for these word pairs.

Given that $\varphi(T)$  is the sum of $M$ independent ad identically distributed variables
divided by $M$, one immediately has
\begin{equation}
{\rm E}\left[\varphi(T) \right] ={\rm E}[\varphi_i(T)],
\,\,\,\,\,\,\,\,\,\,\,\,\,\,\,\,\,\,\,
{\rm V\!ar} \! \left[\varphi(T) \right] = \frac{1}{M} {\rm V\!ar} \! \left[\varphi_i(T) \right],
\label{me3}
\end{equation}
where ${\rm E}[\varphi_i(T)] $ and ${\rm V\!ar} \! \left[\varphi_i(T) \right]$, as already mentioned, 
are computed in Appendix A. It turns out
\begin{equation}
 {\rm E}\left[\varphi(T) \right]  = e^{-2(\mu + \lambda)T} 
\label{me3b}
\end{equation}
and
\begin{equation}
 {\rm V\!ar}\left[\varphi(T) \right] = 
 \frac{1}{M}e^{-2\lambda T}(1-e^{-2\lambda T})e^{-4\mu T} + 
 \frac{1}{ML }e^{-2\lambda T} (1-e^{-2\mu T}) \left[ e^{-2\mu T}+ \frac{1}{N\!\!-\!\!1}  \right].
  \label{me3c}
\end{equation}

As a consequence of equation (\ref{me3b}) 
one has that he effective time from the last common ancestor is
\begin{equation}
T = -\frac{1}{2(\lambda + \mu)} \ln \! \big( {\rm E} \! \left[ \varphi \right] \!  \big),
\label{me4}
\end{equation}
while the observed one  is
\begin{equation}
 \mathcal{T}_{\phi}= -\frac{1}{2(\lambda+\mu)} \ln \! \left(\phi\right).
 \label{me5}
\end{equation}
We end up with the new measure of the relative error on separation time 
\begin{equation}
R_\varphi =  
\frac{1}{4(\lambda + \mu) T} \ln \! 
\left( \frac{ {\rm E} \! \left[ \varphi \right] + 2\sqrt{{\rm V\!ar} \! \left[  \varphi   \right] } }
{ {\rm E} \! \left[ \varphi \right] - 2\sqrt{{\rm V\!ar} \! \left[  \varphi   \right] } } \right),
 \label{me6}
\end{equation}
whose meaning should be clear at this point.

It can be shown that the relative error $R_\varphi$ is always smaller than
the relative error $R_\phi$, a fact that can be also visually appreciated in Fig. 1.
First of all one can write
\begin{equation}
\phi_i(T)=\varphi_i(T)+\chi_i(T),  
\,\,\,\,\,\,\, {\rm where} \,\,\,\,\,\,\,\,\,\,
\chi_i(T) =
\frac{N}{N\!\!-\!\!1} \big(1- \sigma_i(T) \big)\left[ \xi_i (T)  - \frac 1 N \right]
 \label{me7}
\end{equation}
which can be easily verified comparing (\ref{lea1}),  (\ref{lea3}) with  (\ref{me1}).
Moreover, as shown in Appendix A, ${\rm E} \! \left[ \chi_i(T) \right] =0$,
which means ${\rm E} \! \left[ \phi_i(T) \right] ={\rm E} \! \left[ \varphi_i(T) \right] $.
Moreover, always in Appendix A,  it is shown that
${\rm V\!ar} \! \left[  \phi_i(T)  \right] = {\rm V\!ar} \! \left[  \varphi_i(T)   \right] + 
{\rm V\!ar}  \! \left[  \chi_i(T)   \right]$. These equalities equally hold when we
pass to the intensive variables obtained by summing over all concepts and dividing by $M$,
therefore
\begin{equation}
{\rm E} \! \left[ \phi(T) \right] ={\rm E} \! \left[ \varphi(T) \right] ,
\,\,\,\,\,\,\ \,\,\,\,\,\,\,\,\,\,
{\rm V\!ar} \! \left[  \phi(T)  \right] = {\rm V\!ar} \! \left[  \varphi(T)   \right] + 
{\rm V\!ar}  \! \left[  \chi(T)   \right]
 \label{me8}
\end{equation}
which finally imply $R_\varphi \le R_\phi$. The reason of this inequalities 
finally relies on the fact that the contribution of
the Hamming overlaps for non-cognate pairs,
do not add information but only makes the signal more noisy.

At this point one would conclude that one should always use
$\varphi$ instead of $\phi$, but this is not true, because the computation of 
$\varphi$ requests the individuation of all cognate pairs which is
a procedure which can be subject to errors, as explained at the beginning of the paper.

It is true that he relative error $R_\varphi$ is always smaller than $R_\phi$,
but in the first three millennia they are almost equal, therefore the blind strategy  should
to be preferred in this interval of time because the error due to the wrong attribution
of cognacy relations likely produces an extra-variance larger than 
${\rm V\!ar}  \! \left[  \chi(T)   \right]$.

This is still true when using non-subjective automatic strategies for cognacy attribution
\cite{Covington:1996,Ciobanu:2014,List:2017,Rama:2019,Rama:2020,Pasquini:2023},
thus supporting the empirical results in \cite{Rama:2020},
where a comparison was made between the performances of edit-distance dating and glottochronological dating.

We leave the task of comprehensively discussing the advantages and disadvantages of the 
three approaches to the concluding section that follows.

\section{Conclusions}

If we take into account both the results shown in Fig. 1 and the considerations at the end of the 
previous section, it should be quite clear that up to three millennia the most convenient 
strategy is the one in Section 4, which is based only on the use of the normalized edit 
distance for all pairs, thus without the need to identify cognacy relations between words. 
Indeed, both $R_\phi$ and $R_\varphi$ are much smaller than $R_\omega$ and they
are almost equal within this time window, thus between the two strategies based on the normalized 
edit distance the blind one should be preferred because there is no risk of increasing the error as a consequence of a subjective misattribution of cognacy relations.

It is true that we found the values of $N$ and $L$ rather roughly and that smaller actual values 
of $N$ and/or $L$ would increase the variances (\ref{leb1c}) and (\ref{me3c}) 
and, consequently, increase  the error $R_\phi$ and $R_\varphi$.
However, for the first three millennia both $R_\phi$ and $R_\varphi$ 
are almost independent of $N$ and $L$. This is a consequence of the fact that in this time range 
the leading term of the variances (\ref{leb1c}) and (\ref{me3c})
is $ \frac{1}{M}e^{-2\lambda T}(1-e^{-2\lambda T})e^{-4\mu T} $, that does 
not depend on these two parameters.
Indeed, if we consider only this term we find
$$
R_\phi \simeq R_\varphi \simeq  \frac{\lambda}{\lambda+\mu} R_\omega,
$$
which is indeed what can be visually perceived in Fig. 1 for the first three millennia.

For longer time horizons (over three millennia) it appears that the best choice should be the 
mixed strategy, $i.e.$ the one that limits the calculation of the normalized edit distance 
 to cognate pairs (Section 5). 
Only for time horizons longer than six millennia could the Swadesh approach become competitive, 
but six millennia is considered the time when the glottochronological approach tests the limit of its 
validity for reasons inherent in the difficulty of identifying cognacy relations between words.

It must also be said that the approach based on the blind use of the normalized edit distance 
is much more manageable, for example, in the case of the Malagasy language
\cite{Serva:2012a,Serva:2012b,Serva:2020,Serva:2022}, 
60 vocabularies are considered for as many varieties of the language, each with 207 words.
Therefore, cognacy relations would have to be established subjectively for 
$(60 \cdot 59 \cdot 207) /2 = 366,390$ word pairs, a huge task, which, on the contrary, 
can be avoided when using the strategy in Section 4.
Moreover, measuring normalized edit distances can be done authomatically.

Recently there has been some increase in the use of normalized edit distance as a tool for 
investigating also in fields not strictly limited to linguistics, as history, economics, 
migratory phenomena, esthetics and animal communication.
For example, in \cite{Serva:2022} a specific scenario is proposed regarding the modalities of
the Austronesian colonization of Madagascar based on the geographical gradient of introgression 
of Bantu words into the Malagasy varieties.
The influence of language dissimilarity on international economic transactions and on
economic growth are respectively quantified in \cite{Isphording:2013} and \cite{Goren:2018},
while  \cite{Isphording:2014} istudied the effect of linguistic barriers in the destination l
anguage acquisition of immigrants.
More intriguing, the effects of personality traits on phonaesthetic language ratings are investigated 
for the first time in \cite{Winkler:2023} with the conclusion that to some
degree the sound-based beauty of language is “in the ears of the
beholder”, which it is not as trivial as it appears, considered that the study is strictly quantitative.
Finally, there are two different studies carried out by two different research teams that concern the communication between individuals belonging to the whale species Megaptera Novaeangliae
\cite{Magnusdottir:2019, Garland:2012}.

The final conclusion we would like to draw is that methods based on normalized edit distance 
are not only more manageable and easier to use than traditional methods based on cognate 
counting (a fairly accepted fact),  but they are also generally more accurate (a fact which is 
rarely recognized in the linguistic community).  
Moreover, they can be applied much more easily in fields other than traditional linguistics, 
such as animal communication, for which it is difficult to imagine an adaptation of the concept 
of cognacy.

\section*{Acknowledgements}
The research described in this paper has been developed in the framework of the research 
project  National Centre for HPC, Big Data and Quantum Computing - PNRR Project, funded 
by the European Union - Next Generation EU.
\smallskip

The author thanks Michele Pasquini who performed  the automatic calculations in Appendix B,
created the figures and had the patience to discuss some probabilistic aspects of the article.
He also revised the manuscript, but in no way wanted to appear as a co-author.

\section*{Appendix A}
Given (\ref{sc1}) one has  ${\rm E} \left[\xi_{i,k}(T)\right]  =    \frac 1 N  $,
moreover, given that $\xi^2_{i,k}(T)=\xi_{i,k}(T)$ (the variable only takes the values 0 and 1),
one also has ${\rm E} \left[\xi^2_{i,k}(T)\right]  =    \frac 1 N  $ which leads to
${\rm V\!ar}[\xi_{i,k}(T)] =\frac{N\!\!-\!\!1}{N^2} $. 
The variable $\xi_i(T)$ is the normalized Hamming overlap between two non-cognate words
corresponding to the same concept $i$.
Its expression is given by the first equation in  (\ref{sc2}) which is an equal weight average 
of $L$ $i.i.d.$ variables. 
Therefore, one also has 
\begin{equation}
{\rm E} \left[\xi_{i}(T)\right]  =    {\rm E} \left[\xi_{i,k}(T)\right]  = \frac 1 N ,
\,\,\,\,\,\,\,\,\,\,\,\,\,\,
{\rm V\!ar}[\xi_{i}(T)] =  \frac 1 L    {\rm V\!ar}[\xi_{i,k}(T)] = \frac{N\!\!-\!\!1}{LN^2} .
\label{A1}
\end{equation}
This result can be formulated as
\begin{equation}
 {\rm E} \left[ \xi_{i} (T) - \frac 1 N \right]  =  0,
\,\,\,\,\,\,\,\,\,\,\,\,\,\,\,\,\,\,\,
 {\rm E} \left[\left(\xi_{i}(T) - \frac 1 N \right)^2 \right]    =  
 {\rm V\!ar}\left[ \xi_{i} (T)- \frac 1 N \right]  =  {\rm V\!ar}[\xi_{i}(T)]  = 
 \frac 1 L \frac{N\!\!-\!\!1}{N^2} .
\label{A2}
\end{equation}
Let us now define
\begin{equation}
\chi_i(T)=\frac{N}{N\!\!-\!\!1} \big( 1-\sigma_i (T) \big) \left[\xi_{i}(T)- \frac 1 N \right],
\label{A3}
\end{equation}
given that $\sigma_i(T)$ and $\xi_{i}(T)$ are independent variables and  that 
$ {\rm E} \left[\big( 1-\sigma_i (T) \big)^2 \right]  =   {\rm E} \big[ 1-\sigma_i (T) \big]   =
1 - {\rm E} \big[ \sigma_i (T) \big] =1-e^{-2\lambda T}$
and also given the equalities (\ref{A1}), one has

\begin{equation}
{\rm E} \left[  \chi_{i} (T)  \right] =    0,
\,\,\,\,\,\,\,\,\, \,\,\,\,\,\,
{\rm V\!ar} \left [\chi_{i} (T) \right]    =  {\rm E} \left[  \chi^2_{i} (T)  \right] =   
\frac{1}{L(N\!\!-\!\!1)}   (1-e^{-2\lambda T}).
\label{A4}
\end{equation}

We repeat now the above calculations  for variables $\eta_{i,k}(T)$ and related variables.
Given the  probabilities in (\ref{sc5}) and given that
$\eta_{i,k}^2(T)=\eta_{i,k}(T)$, it follows 
\begin{equation}
{\rm E}[\eta_{i,k}^2(T) ]={\rm E}[\eta_{i,k}(T)] = 
\left[ \frac{N\!\!-\!\!1}{N} e^{-2\mu T}+ \frac 1 N \right]
\,\,\, \to \,\,\, 
 {\rm V\!ar}[\eta_{i,k}(T)]  = \frac{N\!\!-\!\!1}{N} (1-e^{-2\mu T}) 
\left[ \frac{N\!\!-\!\!1}{N} e^{-2\mu T}+ \frac 1 N \right].
\label{A5}
\end{equation}
The variable $\eta_i(T)$ is the normalized Hamming overlap between two cognate words
corresponding to the same concept $i$.
Its expression is given by the first equation in  (\ref{sc7}) which is an equal weight average 
of $L$ $i.i.d.$ variables. 
Given that the Bernoulli variables $\eta_{i,k} (T)$ are all independent
the variables $\eta_i(T)$ have the following average value and variance:
\begin{equation}
 {\rm E} \left[\eta_{i}(T) - \frac 1 N \right]  =   {\rm E}\left[ \eta_{i,k} (T) - \frac 1 N \right] 
 =\frac{N\!\!-\!\!1}{N} e^{-2\mu T},
\label{A6}
\end{equation}
\begin{equation}
{\rm V\!ar} \left[\eta_{i}(T) - \frac 1 N \right]      = {\rm V\!ar}[\eta_{i}(T)]  = 
\frac 1 L {\rm V\!ar}[\eta_{i,k}(T)]  =  \frac 1 L \left(\frac{N\!\!-\!\!1}{N} \right)^2 (1-e^{-2\mu T}) 
\left[ e^{-2\mu T}+ \frac{1}{N\!\!-\!\!1}  \right].
\label{A7}
\end{equation}
Notice that (\ref{A6}), (\ref{A7}) reduce to (\ref{A2}) when $T \to \infty$,
$i.e.,$ when the steady state is reached.
Let us consider now the variable $\varphi_i(T)$ defined as
\begin{equation}
\varphi_i(T)=\frac{N}{N\!\!-\!\!1} \sigma_i (T) \left[\eta_{i}(T) - \frac 1 N \right],     
\label{A8}
\end{equation}
given the independence of $\sigma_i (T)$ and $\eta_i (T)$, we have
\begin{equation}
 {\rm E}\left[\varphi_{i}(T) \right]  = e^{-2(\mu + \lambda)T} 
\label{A9}
\end{equation}
and, simply by the definition of variance, we also have
\begin{equation}
{\rm V\!ar}\left[\varphi_{i}(T) \right]  =  \left(\frac{N}{N\!\!-\!\!1} \right)^2 
 \left( {\rm E}\left[\sigma_{i}^2(T)  \right]  {\rm V\!ar}\ \left[\eta_{i}(T) - \frac 1 N \right] +
{\rm E} \left[\eta_{i}(T) - \frac 1 N \right]^2  {\rm V\!ar} \left[\sigma_{i}(T)  \right] \right),
\label{A10}
\end{equation}
Therefore, taking into account that ${\rm E}\left[\sigma_{i}^2(T)  \right] =
{\rm E}\left[\sigma_{i}(T)  \right] = e^{-2\lambda T} $ and 
${\rm V\!ar} \left[\sigma_{i}(T)\right] = e^{-2\lambda T} (1-  e^{-2\lambda T}  )$
we finally obtain
\begin{equation}
 {\rm V\!ar}\left[\varphi_{i}(T) \right] = 
 \frac{1}{L }e^{-2\lambda T} (1-e^{-2\mu T}) \left[ e^{-2\mu T}+ \frac{1}{N\!\!-\!\!1}  \right]
 + e^{-2\lambda T}(1-e^{-2\lambda T})e^{-4\mu T} ,
  \label{A11}
\end{equation}

Let us now define
\begin{equation}
\phi_i(T)=\varphi_i(T) +\chi_i(T)=\frac{N}{N\!\!-\!\!1} \left[\sigma_i (T) \eta_{i}(T) +
\big( 1-\sigma_i (T) \big) \xi_{i} (T)- \frac 1 N \right] ,    
\label{A12}
\end{equation}
given that ${\rm E} \left[  \chi_{i} (T)  \right] =0$, one has
\begin{equation}
{\rm E} \left[  \phi_{i} (T)  \right] ={\rm E}\left[\varphi_{i}(T) \right]  = e^{-2(\mu + \lambda)T} ,
\,\,\,\,\,\,\,\,\,\,\,\,\,\,\,\,\,\,\,\,\,\,
{\rm V\!ar} \left [\phi_{i} (T) \right] = {\rm V\!ar} \left [\varphi_{i} (T) \right] +
{\rm V\!ar} \left [\chi_{i} (T) \right]   
 \label{A13}
\end{equation}
where the last equality is a consequence of the fact that 
$\varphi_{i} (T) \, \chi_{i} (T) =0$,  which in turn is a consequence of the fact that
 $\sigma_i(T) (1- \sigma_i(T) )$ identically vanishes.
 In fact,
\begin{equation}
{\rm V\!ar} \left [\phi_{i} (T) \right] = {\rm E} \left[ \big(  \varphi_{i} (T)  +  \chi_{i} (T) \big)^2\right]
- {\rm E} \left[  \varphi_{i} (T)  +  \chi_{i} (T) \right]^2 = 
{\rm E} \left[ \varphi^2_{i} (T) \right] + {\rm E} \left[ \chi^2_{i} (T) \right] 
- {\rm E} \left[ \varphi_{i} (T) \right]^2
 \label{A14}
 \end{equation}
 where for the second equality we have used ${\rm E} \left[ \chi_{i} (T) \right]=0$.
 Since ${\rm E} \left[ \chi_{i} (T) \right]=0$, it is also true that the last expression above
 equals the last expression in (\ref{A13}). In conclusion:
\begin{equation}
{\rm V\!ar} \left [\phi_{i} (T) \right] =  
\frac{1}{L }e^{-2\lambda T} (1-e^{-2\mu T}) \left[ e^{-2\mu T}+ \frac{1}{N\!\!-\!\!1}  \right]
 + e^{-2\lambda T}(1-e^{-2\lambda T})e^{-4\mu T} + \frac{1}{L(N\!\!-\!\!1)}   (1-e^{-2\lambda T}).
 \label{A15}
\end{equation}
\section*{Appendix B}

The dataset we use in this appendix to estimate the parameters $\lambda$, $\mu$, $N$ and $L$
consists in 60 Swadesh lists of 207 items, overall 12,420 
terms  collected by the author of the present paper during the years 2018 and 2019. 
Each list corresponds to a different variety of Malagasy,  
which is not simply identified by the name of the 
ethnicity but also by the location where the variety was collected. In turn, the
location is identified by the name of a town/village and by latitude and longitude.
This last information turns out to be relevant for results in this appendix since we will 
need of geographical distances for our deductions.

The reasons for using this dataset is that the orthographic realizations are  identical
for all varieties and that the criteria of selection of words is homogeneous and reliable.
In fact, each list was furnished and checked at least by three native language speakers 
which, for each given concept, were asked to furnish the most common word in their 
variety as spoken in their town/village. 
The 60 Swadesh lists dataset, together with the information concerning
ethnicities, towns/villages and latitudes and longitudes, can be found in the 
Supporting Information of \cite{Serva:2020}.

Let us start with parameters $N$ and $L$.
The equalities in (\ref{A2}) can be rewritten in a slightly different form as follows:
\begin{equation}
  \frac 1 N ={\rm E} \left[ \xi_{i} \right] ,
\,\,\,\,\,\,\,\,\,\,\,\,\,\,\,\,\,\,\,
 \frac 1 L  =  \frac{N^2}{N\!\!-\!\!1}  {\rm E} \!\left[\left(\xi_{i} - \frac 1 N \right)^2 \right] . 
 \label{B1}
\end{equation}
were the argument $T$ of the $\xi_i$ has been dropped because inessential
in this context.
The variable $\xi_i$ is the normalized Hamming overlap between two non-cognate words,
while $d_i =1-\xi_i$ is their normalized Hamming distance.
The two above equations can be rewritten using the distances as follows:
\begin{equation}
  \frac 1 N ={\rm E} \left[1- d_i \right] ,
\,\,\,\,\,\,\,\,\,\,\,\,\,\,\,\,\,\,\,
 \frac 1 L  =  (N\!\!-\!\!1) \, {\rm E} \! \left[\left(1 - \frac{N }{N\!\!-\!\!1}d_i\right)^2 \right] . 
 \label{B2}
\end{equation}
It is important to underline that these two equalities hold when the distance 
is between two non-cognate words corresponding to the same concept in two languages, 
but they must also hold when the two compared words correspond to different 
concepts in the same language or  to different concepts in two different languages
(provided they have the same characteristics in terms of effective length and number of letters).

Then, the effective number $N$ of different letter's values can be estimated by 
replacing the theoretical normalized Hamming distance with effectively measurable
normalized Levenshtein distance of words referring to different concepts
\begin{equation}
\frac{1}{N} = \frac{1}{S} \sum_{\alpha, \beta} \sum_{i \neq j} 
\big[1-\mathcal{D}(\alpha,  i \, | \, \beta, j)\big] ,
\label{B3}
\end{equation}
where $\mathcal{D}(\alpha,  i \, | \, \beta, j)$ is the Normalized Levenshtein Distance between 
the word $i$  of language $\alpha$ and the word $j$ of language $\beta$
computed using the dataset collecterd by the author.
The first sum goes on all possible pairs of languages
(included on the same language, $i.e.$, $\alpha = \beta$), 
the second sum only goes on pairs referring to different concepts  ($i \neq j)$.
The normalization is obtained by $S=\sum_{\alpha,  \beta} \sum_{i \neq j} 1$.
The statistical average above replaces the first probabilistic average in (\ref{B2}),
which is reasonable provided the statistics is sufficient.

Following the same procedure we find from the second probabilistic average in (\ref{B2})
that the effective number $L$ of letters in a word can be estimated by
\begin{equation}
\frac{1}{L} = \frac{N \! - \! 1}{S} \sum_{\alpha, \beta} \sum_{i \neq j} 
\left[1 - \frac{N }{N\!\!-\!\!1}\mathcal{D}(\alpha,  i \, | \, \beta, j)\right]^2.
\label{B4}
\end{equation}

Since we consider 60 contemporary varieties of Malagasy, 
the number of languages pairs is  $(60 \cdot 59) /2$ while, given that a list
contains 207 concepts,  the number of pairs corresponding to different concepts
is $207 \cdot 206$ thus, $S=207 \cdot 206 \cdot 60 \cdot 59) /2 \simeq 75 \cdot 10^6$,
a number which is sufficiently large to justify the fact that the statistical averages
(\ref{B3}), (\ref{B4}) reasonably replace the probabilistic ones in  (\ref{B2}).

We find  $N= 5.1770$, and $L=7.6334$. 
There is a big difference between this value of $N$ and the number of letters in the 
Malagasy alphabet which is 21.
This is because $N$ is an effective number that takes into account the fact
that the frequency of the various characters is very different,
moreover, letters are not independent inside words  (syllables,  consonant pairs, \dots).
\bigskip

We perform now an independent evaluation of 
the value of $\lambda$ which will confirm the value estimated by Swadesh.
We rewrite the main formula of glottochronology as
\begin{equation}
\lambda= -\frac{1}{2T } \ln {\rm E} \! \left[  \omega(T) \right] ,
\label{B5}
\end{equation}
then, we consider the 60 contemporary varieties of Malagasy, 
and we replace the probabilistic average (\ref{B5}) 
by the statistical average
\begin{equation}
\lambda(g)=  
-\frac{1}{2T } \ln \left( \frac{1}{R(g)} \sum_{\alpha,\beta} \, \omega (\alpha \, | \,\beta)\right) .
\label{B6}
\end{equation}
where $ \omega (\alpha \, | \,\beta)  $  is the experimentally determined
overlap (number of cognate pairs divided by 207) for varieties $\alpha$ and $\beta$.
The cognates pairs are automatically detected following the procedure in \cite{Pasquini:2019}.
The sum goes on all pairs of varieties which not only match at the root of the tree
(see tree in \cite{Serva:2020}, Fig 2), but also whose 
geographical distance is larger then a given threshold $g$ expressed in $Km$.

The reason why we ask that all pair of languages considered in (\ref{B6})  
must match at the root is that this choice implies that
$T$ in the formula (\ref{B6}) is same for all pairs and it is the time distance
from the last common ancestor of all Malagasy varieties 
(whose known value is $T= 1350 $ years, see \cite{Serva:2012a}).
Moreover, the reason why we ask that all pair of languages considered in (\ref{B6})  
must also be at a geographical distance larger then $g$ is that we would like  to avoid, at 
our best, contamination due to horizontal transfers between geographically close varieties.
The larger is $g$ the smaller is contamination, but unfortunately
the larger is $g$, the smaller is $R(g)$, which is the number of elements in
the average ($R(g) = \sum_{\alpha,\beta} 1$)

The estimated $\lambda(g)$  is plotted in Fig. (\ref{rerr}) (blue) in function of $g$
as well the number of pairs $ R(g) = \sum_{i,j} 1$ considered in the in the average (red).
The larger is $g$, the better is the result because at a larger
geographical distance corresponds a smaller horizontal transfer
between languages. This is true until $R(g)$ becomes too small
for having a sufficient statistics.
In Fig. 2 the geographical distance threshold $g$ ranges between 0 and 1500 $Km$, 
for all values of $g$ the value of $\lambda(g)$ is compatible with the Swadesh estimate
$\lambda =1.4 \times 10^{-4}$, nevertheless, as expected, the best result is 
for $1200 \le g \le 1400$ were the geographical distance is the largest compatible with
the constraint of having a sufficient statistics.

In a similar way we can  estimate $\mu$. Equations  (\ref{lea6})  and (\ref{leb3})
rewrite as
\begin{equation}
\lambda + \mu = -\frac{1}{2T}  \ln   {\rm E} \! \left[ 1- \frac{N}{N\!\!-\!\!1} d^H (T) \right] \! ,
\label{B7}
\end{equation}
which corresponds to the statistical average
\begin{equation}
\mu(g) =  
-\frac{1}{2T } 
\ln \left[\frac{1}{R(g)} \sum_{\alpha,\beta} 
\left( 1 - \frac{N}{N\!\!-\!\!1}\mathcal{D}(\alpha \, | \, \beta)\right) \right]
-\lambda(g),
\label{B8}
\end{equation}
where $\mathcal{D}(\alpha \, | \, \beta)$ is the ordinary normalized Levenshtein distance
between the languages $\alpha$ and $\beta$ (of course, $\alpha \neq \beta$) and where,
again, the sum goes on all pairs of varieties which not only match at the root 
of the tree (see the tree in \cite{Serva:2020}, Fig 2), but also whose 
geographical distance is larger then a given threshold $g$ expressed in $Km$.

In Fig. 2, indeed, we plot $\hat \mu(g)$ (green) instead of $\mu(g)$ because it is the rate of
an effective character change. 
The most reliable result is  for $1200 \le g \le 1400$ were the geographical distance is the 
largest compatible with the constraint of having a sufficient statistics.
We find $\hat \mu =1.3 \times 10^{-4}$,
very close to the value $\lambda =1.4 \times 10^{-4}$ which means that character changes
and word replacements equally contribute to the lexicon evolution.
\begin{figure}
\centering
\includegraphics[height=9cm,width=15cm,angle=0]{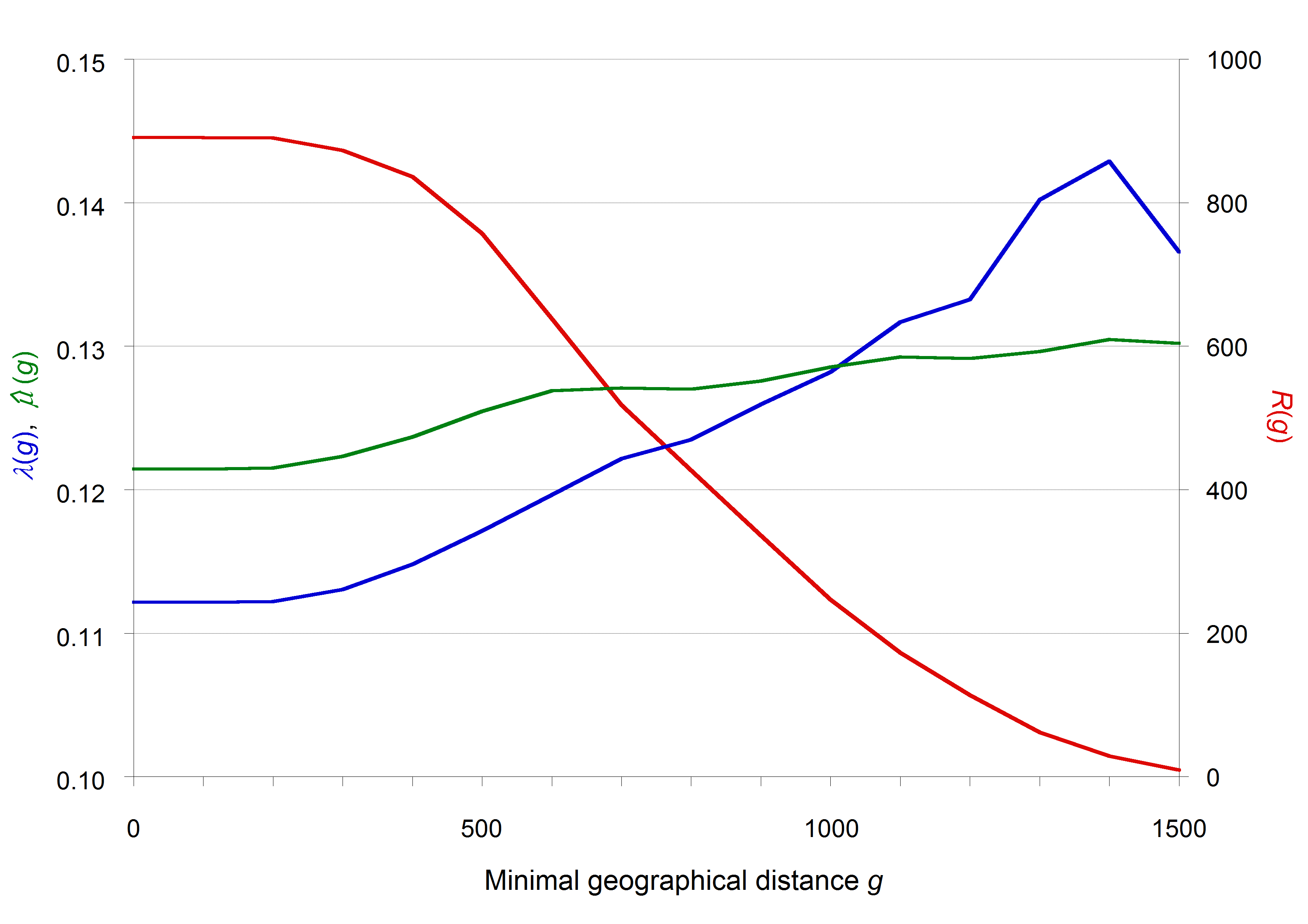}
\caption{The parameters $\lambda(g)$ (blue), $\hat \mu(g) =  \frac{N\!\!-\!\!1}{N} \mu(g)$ (green)
and $R(g)$ (red) as a function of the minimal geographical distance $g$ (in $Km$) between two 
varieties in a pair. The parameter $\lambda(g)$ is the rate of words replacements,
$\hat \mu(g)$ is the rate of  effective characters changes and $R(g)$ is the number of 
language pairs involved  in the average, $i.e.,$ the number of language pairs which 
match at the root of the family tree and whose geographical distance is larger than $g$.}
\label{rate}
\end{figure} 

\end{document}